\documentclass[10pt,twocolumn,letterpaper]{article}

\usepackage[pagenumbers]{wacv} %

\usepackage{graphicx}
\usepackage{booktabs}
\usepackage{url}
\usepackage{algorithm}
\usepackage{algorithmic}
\usepackage{amsmath,amssymb}
\usepackage[normalem]{ulem}
\usepackage{soul}
\usepackage{xcolor}

\usepackage[accsupp]{axessibility}  %

 \def\vs{\emph{vs}\onedot}
\def\wrt{w.r.t\onedot} 

\makeatletter
\DeclareRobustCommand\onedot{\futurelet\@let@token\@onedot}
\def\@onedot{\ifx\@let@token.\else.\null\fi\xspace}
\makeatother

\definecolor{wacvblue}{rgb}{0.21,0.49,0.74}
\usepackage[pagebackref,breaklinks,colorlinks,allcolors=wacvblue]{hyperref}

\def\wacvPaperID{1172} %
\def\confName{WACV}
\def\confYear{2027}

\begin{document}

\title{Selective Timestep Weighting and Advantage-Based Replay for Sample-Efficient Diffusion RLHF}

\author{Eric Zhu\\
Carnegie Mellon University\\
{\tt\small ezhu3@andrew.cmu.edu}
\and
Abhinav Shrivastava\\
University of Maryland, College Park\\
{\tt\small abhinav2@umd.edu}
\and
Soumik Mukhopadhyay\\
University of Maryland, College Park\\
{\tt\small soumik@umd.edu}
}

\maketitle

\begin{abstract}

\vspace{-0.5em}
    Reinforcement learning from human feedback (RLHF) has emerged as a powerful paradigm for aligning generative models with human preferences. However, applying RLHF to diffusion models remains highly feedback inefficient, as existing approaches typically require large amounts of human or reward model evaluations. This limitation reduces the practicality of diffusion RLHF in real-world settings where feedback is the primary bottleneck.  In this paper, we propose two complementary strategies that substantially improve the feedback efficiency of diffusion RLHF while preserving generalization to unseen prompts. Our key observation is that reward information in diffusion trajectories is unevenly distributed: not all denoising timesteps or trajectories contribute equally to learning from a reward signal. By emphasizing informative timesteps and trajectories during optimization, we obtain more effective gradient updates. First, we introduce a per-timestep weighting scheme that reweights denoising steps during policy optimization. We theoretically connect this weighting to the optimal convergence properties of proximal policy optimization (PPO) and approximate the resulting weighting trend empirically. Second, we introduce a replay mechanism that prioritizes informative trajectories, enabling the model to reuse past samples instead of repeatedly querying new rewards.  Together, these strategies significantly improve the feedback efficiency of diffusion RLHF. Under identical hyperparameter settings, our approach achieves up to a $6\times$ improvement in sample efficiency compared to widely used diffusion RLHF baselines.
\end{abstract}

\section{Introduction}

Diffusion models~\cite{OriginalDiffusionPaper} have become the leading framework for high-quality image generation. However, because they are trained to reproduce the distribution of their training data, they do not inherently reflect human preferences. Recent work~\cite{DDPO, DPOK} addresses this limitation through reinforcement learning from human feedback (RLHF)~\cite{Original_RLHF_paper}, which fine-tunes diffusion models using scalar feedback from human or reward models to explicitly optimize for preference alignment.
\begin{figure}[t]
    \centering
    \includegraphics[width=0.7\linewidth]{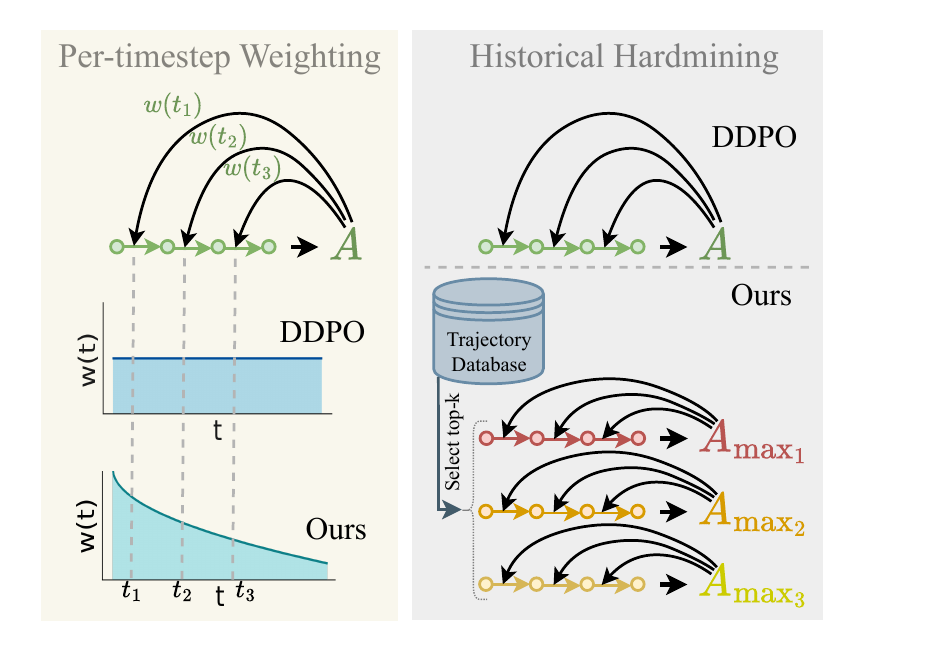}
    \vspace{-0.5cm}
    \caption{Our method has two parts. In the per-timestep weighting, we consider the relative importance of each transition step in the denoising trajectory. In the historical hardmining part, we look at previous trajectories with high advantage and repeat those trajectories in the training.}
    \label{fig:method}
    \vspace{-0.2cm}
\end{figure}
\begin{figure}[t]
    \centering
    \includegraphics[width=1.0\linewidth]{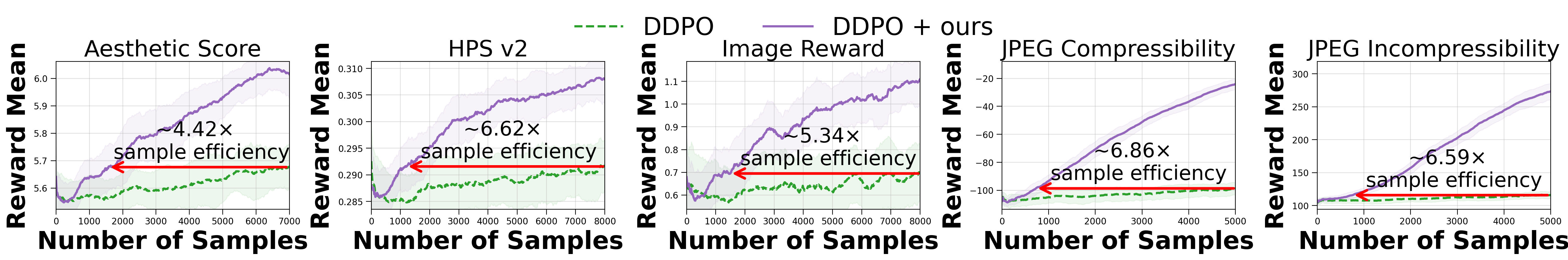}
    \vspace{-0.6cm}
    \caption{
    A graph of a baseline method (bottom curve) \vs the same baseline method using our method (top curve). As shown above, augmenting existing RLHF frameworks with our method achieves significant sample efficiency compared to the baseline.
    }
    \label{fig:placeholder}
    \vspace{-0.5cm}
\end{figure}

One major challenge with RLHF in diffusion models is the credit assignment problem: the difficulty of determining which intermediate timesteps in the diffusion process actually contributed to the final reward. Because human feedback is only given on the final image, methods such as DDPO~\cite{DDPO} simply assign the same loss to every timestep. This setup ignores the structure of the denoising process and how different timesteps edit the image at different levels of granularity~\cite{mukhopadhyay2024textfreediffusionmodelslearn}. As a result, the model optimizes transitions uniformly without considering the non-uniform nature of the denoising trajectory, leading to inefficient training.

Prior work attempts to address this issue by contrasting paired denoising trajectories generated from the same initial noise~\cite{B2-DiffuRL, d3po}. In these approaches, trajectories remain identical until a designated branching timestep and diverge afterward. While contrasting two trajectories with shared initial noise can highlight the effect of the timestep at which they diverge, the approach does not contrast other timesteps in the sequence. Earlier steps remain identical and therefore uninformative, and later steps still inherit the same uniform advantage, leaving the broader credit assignment problem unresolved. Moreover, these methods require multiple reward evaluations per starting noise, reducing sample efficiency and increasing computational cost.

In this paper, we argue that reward information in diffusion trajectories is inherently unevenly distributed across timesteps and trajectories. Some denoising steps and trajectories contribute significantly more learning signal than others. To exploit this observation, we propose two complementary strategies that emphasize informative training data.

First, we introduce a per-timestep weighting scheme that selectively reweights denoising steps during optimization. We theoretically motivate this weighting by analyzing the relationship between Group Relative Policy Optimization (GRPO) and Proximal Policy Optimization (PPO), showing that each timestep should ideally be weighted proportionally to the variance of the TD-error PPO advantage. Although this quantity is infeasible to estimate efficiently during training, we analyze its behavior and approximate the trend using a simple heuristic based on the squared magnitude of per-timestep latent change. Empirically, this weighting improves training sample efficiency.

Second, we introduce a trajectory replay mechanism inspired by replay buffers commonly used in robotics-based reinforcement learning~\cite{prioritizedexperiencereplay, HindsightExperienceReplay, SyntheticER}. Instead of discarding previously sampled trajectories, we reuse informative trajectories during training. Drawing inspiration from prioritized replay buffers~\cite{prioritizedexperiencereplay}, we find that hard-mining trajectories with the largest advantages~\cite{AbhianvHardminingPaper} provides the most sample-efficient learning. By emphasizing informative trajectories, the model can continue improving without repeatedly sampling new trajectories and reward evaluations.

Overall, our method improves the efficiency of the baseline by 2--6$\times$ while remaining simple and compatible with existing diffusion RLHF pipelines.

In summary, our main contributions are:
\begin{itemize}
\item A computationally practical and mathematically motivated per-timestep weighting scheme that mitigates the credit assignment problem in diffusion RLHF.
\item A replay buffer mechanism that retrieves informative past trajectories during training, reducing the need for repeated reward queries.
\item A simple, plug-and-play method that integrates seamlessly into existing diffusion RLHF pipelines without architectural changes.
\item Extensive experiments demonstrating consistent gains across diverse reward functions, underscoring the generality of our approach.
\end{itemize}

\section{Related Work}
\subsection{Replay Buffers}
A replay buffer stores previous environment interactions for a policy to train on. It is standard in off-policy methods such as DDPG~\cite{DDPG}, Soft-Actor Critic~\cite{SAC}, and Deep Q-Networks~\cite{DQN}, but not in on-policy methods like PPO~\cite{PPO} and TRPO~\cite{TRPO}, or in current RLHF methods, since these rely on sampling from the most recent weights to stay in-distribution.

Many replay-buffer variants exist. Hindsight experience replay~\cite{HindsightExperienceReplay} relabels failed trajectories with new goals to synthesize successful ones. Others prioritize transitions under certain criteria.  The approach \cite{prioritizedexperiencereplay} samples by high temporal-difference (TD) error as a proxy for how undertrained a network is at a given point, and Energy-Based Hindsight~\cite{EnergyBasedHindsight} prioritizes high-kinetic-energy trajectories. Synthetic trajectories generated by diffusion models were proposed by~\cite{SyntheticER}, and retrieval was shown by~\cite{RoboticImportanceWeightedRetrieval} to enhance robot imitation learning. Finally,~\cite{liang2021ptrppoproximalpolicyoptimization} hardmines previous PPO trajectories using cumulative TD error.

\subsection{RLHF}
RLHF has become the central paradigm for aligning large language models (LLMs). Modern RLHF was introduced by~\cite{ouyang2022training}, who scaled it to pretrained LLMs via a three-step pipeline of supervised fine-tuning, reward modeling, and reinforcement learning. Subsequent approaches such as Group Relative Policy Optimization~\cite{GRPO} remove the reward model entirely in favor of group-normalized advantage, and~\cite{RL-Razor} shows GRPO prevents catastrophic forgetting better than supervised fine-tuning. Preference-based approaches such as~\cite{DPO_LLM} have a human rank one sample over another rather than give a numerical reward. More recently,~\cite{MA_RLHF} segments the LLM token stream before computing segment-level reward, and~\cite{Hardming-Prompts-LLM-GRPO} applies GRPO with prompt hardmining.

RLHF has also been explored in robotics:~\cite{tri-trend, robotic-rl-vlm} train policies on preference-based evaluation of robot trajectories,~\cite{Diffusion-ES} uses reward-guided evolutionary search over a pretrained trajectory diffusion model, and~\cite{DPPO} finetunes pretrained imitation-learning diffusion policies such as diffusion policy~\cite{DiffusionPolicy}.

\subsection{Diffusion RLHF}
RLHF is also effective for aligning diffusion models with human preference. Diffusion RLHF began with DPOK and DDPO~\cite{DPOK,DDPO}, which applied RL to diffusion and outperformed reward-weighted regression (RWR)~\cite{RWR}. Later approaches such as B2-DiffuRL~\cite{B2-DiffuRL} and Branch-GRPO \cite{branch-grpo} branch from shared input noise to sharpen contrast between trajectories, and another line of work demonstrates GRPO can be applied in flow matching and diffusion~\cite{Flow-GRPO, DanceGRPO, Finegrained-GRPO, branch-grpo}. In contrast to GRPO, preference-based methods~\cite{d3po, dspo, Diffusion-DPO} have a human pick between generated image pairs rather than assign a numerical reward and train on this feedback. Diffusion RLHF has been shown effective in video diffusion~\cite{VADER} and 2-step diffusion~\cite{Two-Step-Diff-RLHF} settings.

In contrast, some methods~\cite{draft, luo2025dualprocessimagegeneration, focusnfix} assume a differentiable reward (e.g.\ a classifier or VLM) and backpropagate through it directly, while others localize spatial regions of reward importance~\cite{PatchDPO, focusnfix}. The work SIPO~\cite{ImportanceTimestepSDPO} reweights trajectory timesteps for more stable training under non-numerical pair-wise preference feedback, and 
a recent work TempFlow~\cite{he2025tempflow} explores timestep weighting in flow matching rather than DDPM-style models, but without hardmining. Moreover, they evaluate on a single reward function, whereas our evaluation across 5 reward functions shows their weighting scheme underperforms ours.

\section{Preliminaries}
\subsection{Markov Decision Process}
Formally, a MDP ~\cite{Intro_To_RL} can be represented as a tuple $(S, A, P, R,\gamma)$, where $S$ is the state space, $A$ is the action space, $P(s' | s,a)$ is the probability transition function from current state $s$ and action $a$ to new state $s'$, $R(s,a)$ is the reward function, and $\gamma$ is the discount factor. At each timestep $t$, the agent has access to state $s$ and chooses an action $a$. The result is a new state $s'$ sampled from $\mathcal{P}(s'| s,a)$ and a reward, denoted as $R(s,a)$.

The goal in reinforcement learning is to maximize the sum of decayed rewards, formally denoted as
$J(\pi) = \mathbb{E}_{\tau\sim p_\pi}\left[\sum_{i=0}^n\gamma^iR(s_i,a_i)\right]$.

\subsection{Diffusion Models}
For a given data sample $x_0 \in \mathbb{R}^d$, the forward diffusion process gradually adds noise over $T$ steps to attain the noised trajectory $(x_0, x_1, ..., x_T)$. Mathematically, the forward process can be defined as
$q(x_t | x_{t-1}) = \mathcal{N}(x_t ; \sqrt{1 -\beta_t}x_{t-1}; \beta_tI)$,
where $\beta_t$ is the monotonic noise schedule. The reverse diffusion process consists of a trained model $p_\theta$ where $p_\theta(x_{t-1} | x_t)$ learns to approximate the posterior distribution $q(x_{t-1}| x_t, x_0)$, which is also a normal distribution with monotonic variance schedule $\tilde \beta_t$.

With reward feedback, the reverse diffusion process can be interpreted as a MDP with the action being the predicted noise from the diffusion model, the state being the latent space of images, and the transition function being a sampler such as DDIM ~\cite{DDIM} and DDPM ~\cite{DDPM} that inputs the current latent and predicted noise.

\subsection{PPO and Diffusion RLHF}
PPO~\cite{PPO} is a commonly used method in classical reinforcement learning. Formally, the PPO uses the loss function: 
$$L_{\text{PPO}} = \min\big(
A_t \tfrac{\pi_\theta(a|s)}{\pi_{\theta_{\text{old}}}(a|s)},
\operatorname{clip}(\tfrac{\pi_\theta(a|s)}{\pi_{\theta_{\text{old}}}(a|s)},1-\epsilon,1+\epsilon)A_t
\big)$$
where $A_t$ is the advantage at timestep $t$, $\epsilon$ is the clipping value to prevent overly large weight shifts, and $\pi_{\theta_\text{old}}(a|s)$ is the probability of taking action $a$ with the old parameters in $\theta_\text{old}$ as opposed to the current parameters $\theta$. In classical robotics RL, the advantage is given by the TD-error 
$A_t = \gamma V(s_{t+1}) + R_{t+1} - V(s_t)$,
where $V$ is the value function. In the context of diffusion RLHF, prior methods do not use the TD-error but rather compute the uniform advantage $A=\frac{r - r_\text{mean}}{\text{std}(r)}$, where $r$ is the reward at the end of the trajectory, $r_\text{mean}$ is the average reward in a batch of samples, and $\text{std}(r)$ is the standard deviation of the rewards in a batch. The same advantage value $A$ is used for all timesteps in the trajectory.

\begin{figure}[t]
    \centering
    \includegraphics[width=\linewidth]{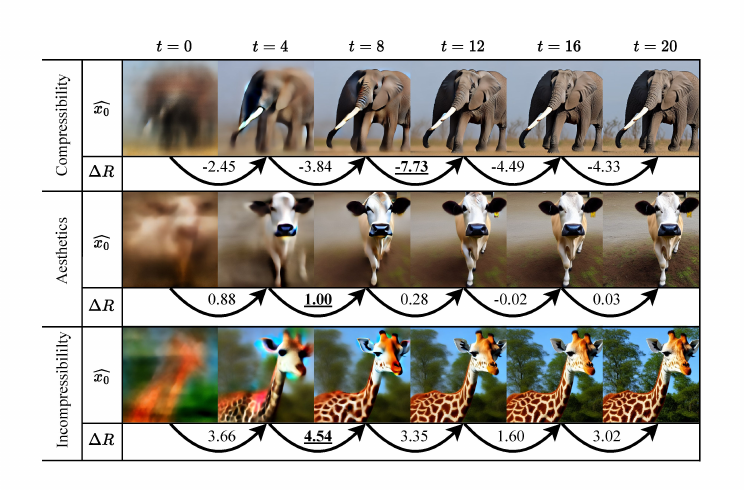}
    \vspace{-0.5cm}
    \caption{
    Change in reward of predicted $x_0$ from step 0 to 4, 4 to 8, and so on.
    This can be used as a proxy for the effect of different timesteps $t$ on the final reward.
    Interestingly, most of the details in the final image are determined by step 12.
    }
    \label{fig:trajectory_change_in_reward}
\end{figure}

\begin{figure}[t]
    \centering
    \includegraphics[width=0.7\linewidth]{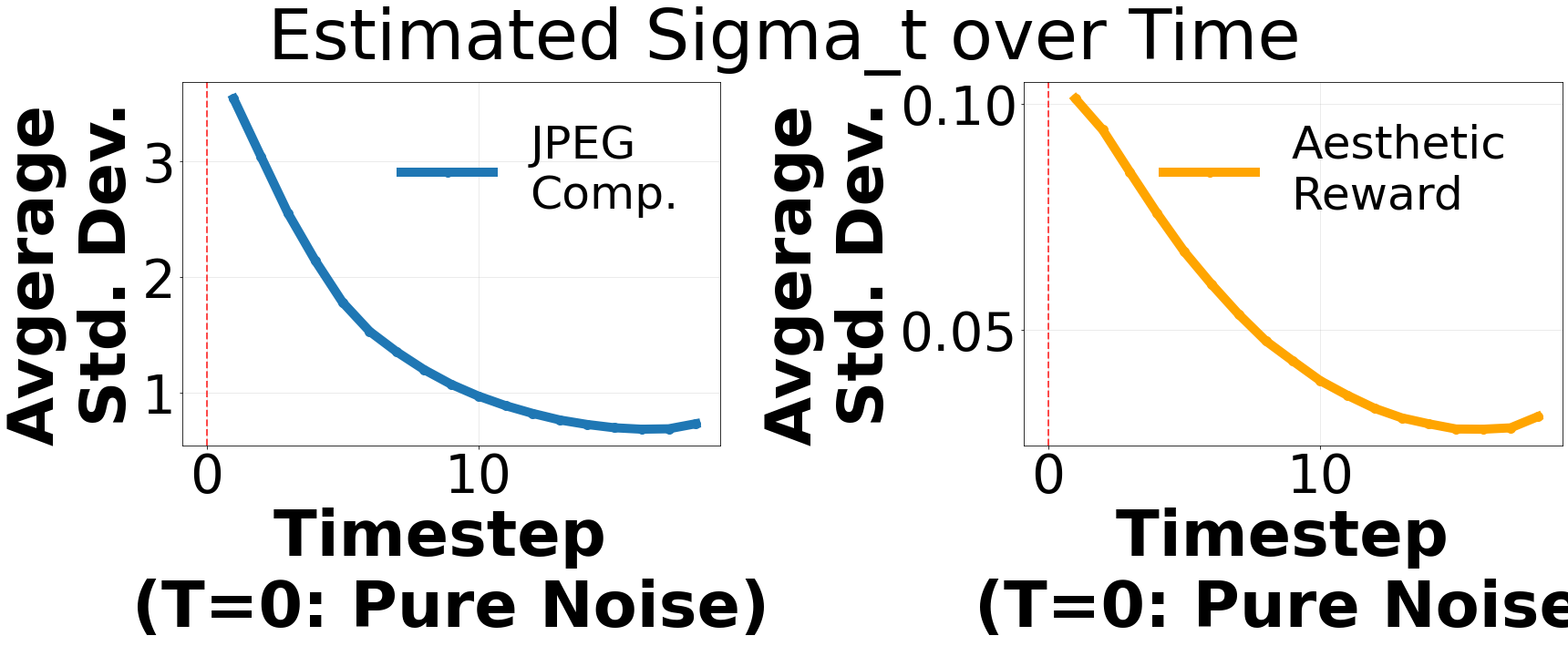}
    \vspace{-0.2cm}
    \caption{Stochastic estimation of $\sigma_t=\sqrt{\text{Var}(A_t)}$.}
    \label{fig:estimate_advantage}
    \vspace{-0.3cm}
\end{figure}

\begin{figure}[t]
    \centering
    \includegraphics[width=0.75\linewidth]{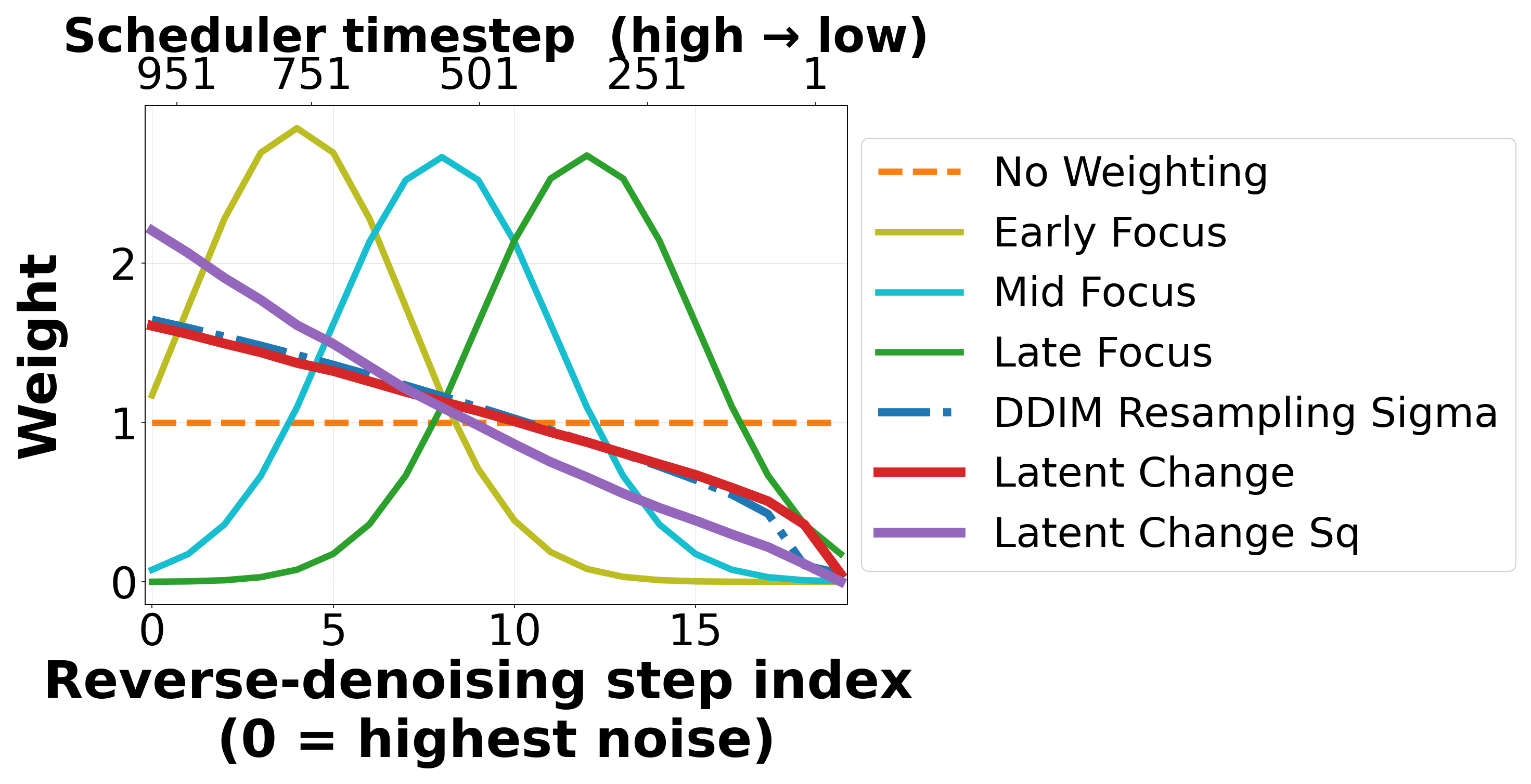}
    \vspace{-0.2cm}
    \caption{Various timestep weighting schemes. For fairness, all timestep weightings are normalized to have a mean of 1.}
    \label{fig:weight_choices}
    \vspace{-0.3cm}
\end{figure}

\section{Methodology}
\begin{figure*}[t]
    \centering
    \includegraphics[width=0.75\linewidth]{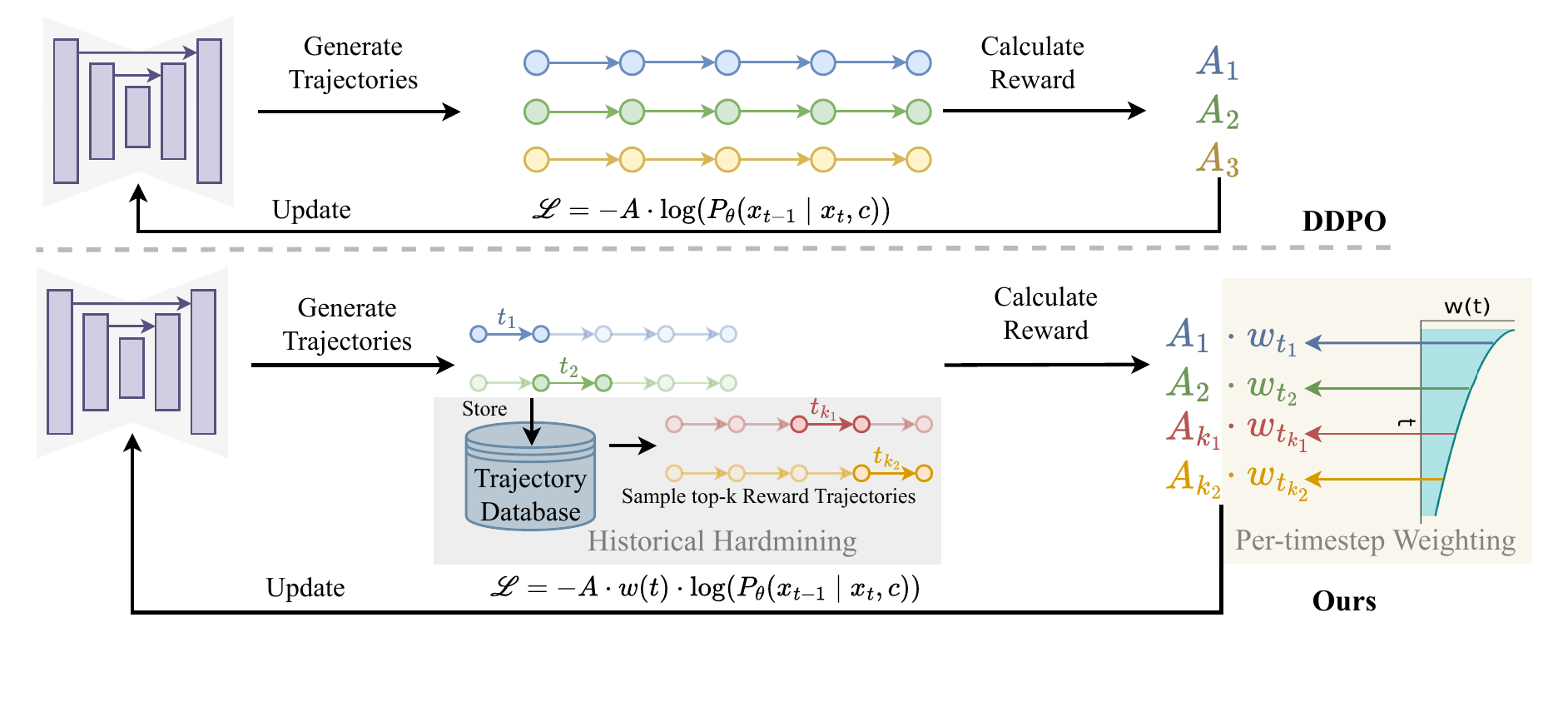}
    \vspace{-0.7cm}
    \caption{An overview of our method. Compared to DDPO, we add a timestep-dependent weight that accounts for the asymmetric nature of the denoising process and add a replay buffer for storing previous trajectories and hardmining the important ones.}
    \label{fig:teaser}
    \vspace{-0.3cm}
\end{figure*}
In existing diffusion RLHF pipelines, informative training signals are often obscured by uninformative or low-reward data. 
Our approach addresses this issue at two levels: (1) timestep-level credit assignment and (2) trajectory-level data selection. We first analyze the credit assignment problem and derive a timestep-dependent weighting scheme, and then introduce a replay-based hard-mining strategy to prioritize informative trajectories.

First, at the timestep level (Sec.~\ref{subsec:strat1}), 
we try to address the credit assignment problem.
While standard diffusion RLHF methods assign uniform credit to all denoising steps, we introduce a practical approximation to a theoretically motivated weighting scheme that assigns greater importance to diffusion steps that contribute most to the final reward
(Fig.~\ref{fig:method} (left)). 
Second, at the trajectory level (Sec.~\ref{subsec:strat2}), we introduce a hard-mining mechanism that selectively reuses past trajectories with high absolute advantage, ensuring that the most informative experiences are emphasized during learning (Fig.~\ref{fig:method} (right)).

\subsection{Non-Uniform Credit Assignment}
\label{subsec:non_uniform_credit_math}
Traditional diffusion RLHF uses GRPO math to compute a single advantage $A = \frac{R - R_\text{mean}}{\text{std}(R)}$. Depending on the diffusion RLHF method, $R_\text{mean}$ and $R_\text{std}$ either refer to the group or batch mean and standard deviation. This advantage constant $A$ is applied evenly in its loss function. The same constant is used in each timestep: 
$$\mathcal{L}_{\text{ddpo}} = \sum_{t=1}^n -A \cdot\log\left(P_\theta(x_{t-1} | x_t, c)\right)$$
This loss function is problematic because it does not account for how much each step contributes to the final reward. 
To better understand the credit assignment issue, we first analyze how reward evolves along a diffusion trajectory.

\subsubsection{Case Study.}
Let's look at the reward corresponding to the predicted clean image $\hat{x}_0(x_t)$ at timestep $t$ be defined as $R_t = R(\hat{x}_0(x_t))$ for some reward function $R$. For brevity, we denote this as $R(x_t)$ since $\hat{x}_0$ varies smoothly with $x_t$. Consider the quantity $\Delta R_t = R(x_{t+1}) - R(x_t)$. For a small diffusion update $x_{t+1} = x_t + h_t$, a first order Taylor series expansion gives $R(x_{t+1}) \approx R(x_t)+h_t^\top\nabla R(x_t)$. Summing over timesteps ($t$) yields a telescoping decomposition $R(x_T)-R(x_0)\approx\sum_t \Delta R(x_t)$ under the assumption of local linearity. This provides an approximate decomposition of the final reward in terms of $\Delta R_t$ and initial reward. If we disregard the initial reward $R(x_0)$ (which should average to the same quantity given random initial noise), $\Delta R_t$ can be viewed as a crude proxy for the contribution of timestep $t$ to the final reward. In Fig.~\ref{fig:trajectory_change_in_reward}, we observe that $\Delta R_t$ varies significantly across timesteps, suggesting that different parts of the denoising trajectory contribute unequally to the final reward signal.  We emphasize that $\Delta R_t$ should not be used directly for credit assignment, since it captures only the immediate reward change and does not account for the long-term effect of the state $x_t$ on future denoising steps.

\subsubsection{Advantage Variance Proportional Weighting as a Natural Consequence.}
Motivated by this observation, we next connect diffusion RLHF to PPO to derive a principled timestep-level credit assignment scheme.

A key difference between PPO and GRPO is that PPO does per-timestep advantages based on TD-error, thereby reducing the credit assignment issue that occurs with GRPO. We derive an equation to reconstruct the PPO timestep-level advantages from a singular final advantage that is computed using GRPO. We show that the PPO timestep advantages are proportional to the trajectory level advantage $A_\text{final}$ for some timestep-dependent constants $w(t)$. By using these PPO advantage values over the uniform GRPO advantages, we can attain better convergence.

The original PPO uses TD-error advantages as: 
$$A_t = \gamma V(s_{t+1}) + R_{t+1} - V(s_t)$$
where $A_t$ is the timestep-level advantage and $V$ is the value function from reinforcement learning. 

For simplicity of notation, we assume a 20 step diffusion process. Let state $s_k = \{x_k, T_k\}$ where $T_k$ is the diffusion timestep and $x_k$ is the noisy image. Consider a diffusion trajectory $(s_{0}, s_1, ..., s_{20})$ where state $s_0$ corresponds to pure noise, $s_{20}$ corresponds to the final image, and $s_k$ corresponds to intermediate latents. Consider the series of actions $a_0, ..., a_{19}$. In a RLHF scenario, there is only a single reward from $s_{20}$, so denote the final reward as $R_\text{final}$ and all intermediate rewards as 0. Finally, we assume that $\gamma=1$, as is typical in RLHF.

We have that for $t = 0, ..., 19$ our equation for $A_t$ simplifies to:
$$A_t = \gamma V(s_{t+1}) + R_{t+1} - V(s_t) = V(s_{t+1}) - V(s_t)$$
We have that $V(s_{20}) = R_\text{final}$. This leads us with 
$$A_0 + A_1 +... + A_{19} $$
$$= \left[V(s_1)-V(s_0)\right] + \left[V(s_2)-V(s_1)\right] + ... + \left[V(s_{20})-V(s_{19})\right] $$
$$= V(s_{20}) - V(s_0)= R_\text{final} - V(s_0)$$

Since $s_0$ is pure noise sampled from $\mathcal{N}(0,1)$, there is a very high diversity of denoising trajectory that can originate from the same start noise due to the high resampling noise of existing samplers. Thus, we assume that $V(s_0) \approx \mathbb{E}_{s_0\sim N(0,1)}(R_\text{final}) = R_\text{mean}$. Thus,
$$A_0+... + A_{19} = R_\text{final} - V(s_0) \approx R_\text{final} - R_\text{mean} = A_\text{final}\cdot \text{std}(R)$$
Under the assumption that we only know the final advantage $A_\text{final}$ and nothing else, we can estimate each $A_t$.
We have that $A_t = V(s_{t+1})-V(s_t)$ is zero-meaned over all trajectories because 
$$V(s_{t}) = \mathbb{E}_{s_{t+1}|s_t}\left[\gamma V(s_{t+1}) + R_t\right] = \mathbb{E}_{s_{t+1}|s_t}\left[V(s_{t+1})\right]$$
$$\implies \mathbb{E}_{s_{t+1}|s_t}\left[A_t\right] = \mathbb{E}_{s_{t+1}|s_t}\left[V(s_{t+1})-V(s_{t})\right] = 0$$

Let $\sigma_t$ be the standard deviation of $A_t$ over the distribution of trajectories generated by $\pi$. We further approximate each $A_t$ as coming from a gaussian $\mathcal{N}(0,\sigma_t^2)$.

We want to find the conditional distribution for   $P\!\left(A_k \mid \sum_{i=0}^{19}\!A_i = A_{\text{final}}\!\cdot\!\operatorname{std}(R)\right)$. This is the conditional distribution of a multivariate normal under linear constraint. This gives us the distribution: 

$$A_k \Big | \sum_{i=0}^{19}A_i = A_\text{final} \cdot \text{std(R)}
  \sim \mathcal{N}\!\left(\mu, \sigma^2\right),$$
$$\text{where } 
\mu = \frac{\sigma_k^2 \, \text{std}(R)}{\sum_{i=0}^n \sigma_i^2}\, A_\text{final},\quad
\sigma^2 = \sigma_k^2\!\left(1 - \frac{\sigma_k^2}{\sum_{i=0}^n \sigma_i^2}\right)$$

Refer to appendix for more details on how this distribution was derived. Thus, the expected value/mean of $A_k$ is a constant times $A_\text{final}$ where the constant is: 
$$w(t) = \frac{\sigma_t^2\cdot \text{std}(R)}{\sum_{i=0}^n \sigma_i^2}$$
Note that $w(t)$ is larger when $\sigma_k$ is larger, which happens in timesteps that change the final reward the most. Thus, if we align a schedule $w(t)$ correctly with $\sigma_t$, then we can construct accurate estimates of the PPO timestep-level advantages $A_t$. In our formulation, $\sigma_t$ does not depend on $s_t$ or $a_t$ and solely depends on the timestep $t$. This allows us to use the same $w(t)$ universally for different trajectories.

\subsubsection{Empirical Measurement of $\text{Var}(A_t)$}
Intuitively, the advantage variance $\text{Var}(A_t)=\sigma_t^2$ should vary across diffusion timesteps, since the forward process adds noise with timestep-dependent variance $\beta_t$. In Fig.~\ref{fig:estimate_advantage} we estimate $\sigma_t^2$  by denoising a trajectory partially till timestep $t$, and branching off into multiple denoising trajectories and computing variance of their final rewards.
 This is undesirable during training due to the need of large number of branches needed to accurately compute the variance. However, this clearly indicates the need for a non-uniform weighting scheme and provides an initial indication of the expected weighting trend. We see here a monotonically decreasing trend (except for a slight increase in the final steps, which might be due to image generation stochasticity).

\begin{figure}[t]
    \centering
    \includegraphics[width=\linewidth]{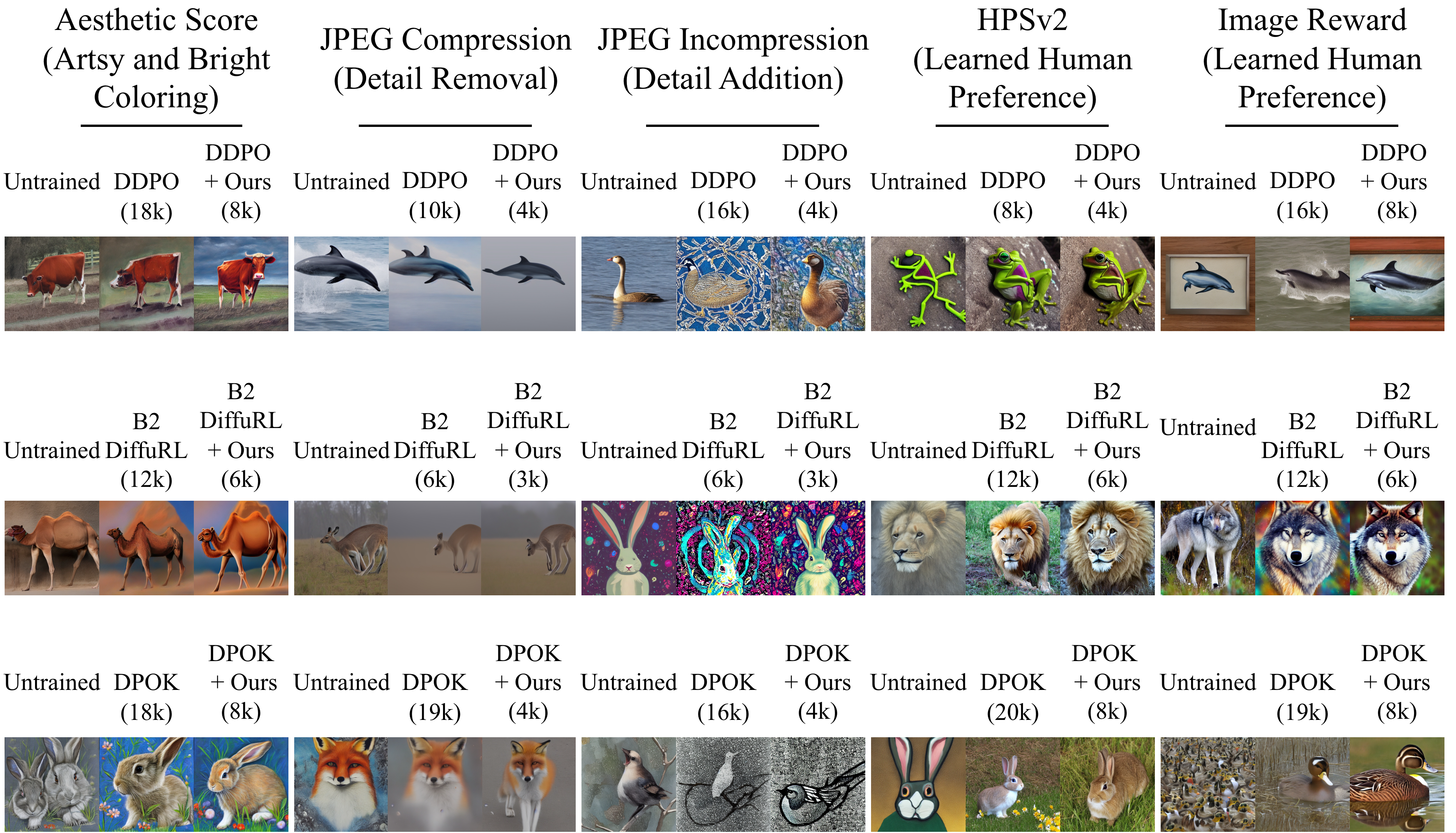}
    \caption{Qualitative comparisons between our method \vs default RLHF method \vs no training. In parentheses is how many reward queries were used during finetuning. All images within triplet are rendered using the same input seed and noise. }
    \label{fig:triplet_qualitative}
    \vspace{-0.4cm}
\end{figure}

\subsection{Strategy 1: Per-Timestep Weighting}
\label{subsec:strat1}

Previously, we motivated the case for using a non-uniform weighting scheme using a case study. 
We then demonstrated that the PPO timestep-level advantage can be derived from the original GRPO advantage under certain assumptions, and that it depends directly on the variance of the advantage. 
While this derivation provides theoretical guidance, estimating these quantities exactly during training is impractical. We therefore seek a simple, practical weighting scheme that approximates this behavior.

We experiment with multiple weighting schemes as shown in Fig.~\ref{fig:weight_choices}.
An obvious choice for such a monotonically decreasing weighting scheme is using the reverse diffusion standard deviation $\sqrt{\tilde\beta_t}$ used for adding randomness during sampling. (A 
recent
work~\cite{he2025tempflow} tried to use a similar strategy of weighting using the standard deviation of noise in Flow-GRPO SDE.) 
We additionally consider the change in diffusion model's mean absolute latent change $|z_{t}-z_{t-1}|$ and mean squared latent change $|z_{t}-z_{t-1}|^2$ as other choices for weighting schemes, which share the same monotonicity. Here, $z_t$ is the diffusion state in the latent space of Stable Diffusion VAE. As opposed to the previous static weighing schemes, this last scheme is dynamic, as it is dependent on the latents which in turn are dependent on how the current model denoises. 
Finally, we also try gaussian weighting schemes with early, middle, and late focus \wrt to the sampling steps. These are not monotonic but can signal towards which sampling step should be focused on more.

We replace the GRPO advantage $A_\text{final}$ with our new advantage $w(t) A_\text{final}$. Our new loss is:
\begin{equation*}\mathcal{L}_{\text{reweighted}} =  \sum_{t=1}^n -A_\text{final} \cdot w(t)\cdot\log\left(P_\theta(x_{t-1} | x_t, c)\right)\end{equation*}
where $w(t)$ is a scalar that only depends on the timestep of the transition. 

\subsection{Replay Buffers}
So far, we focused on improving credit assignment at the timestep level. We now address the second source of inefficiency: how trajectories are reused during training. In classical RL for robotics, methods~\cite{DQN, DDPG, SAC} often make use of a replay buffer that stores previous trajectory from previous epochs. During training, episodes are drawn from the replay buffer and used to train the model alongside data from the current epoch. This allows the policy to sample from a larger dataset than just the trajectories from the current epoch. An extension of replay buffer is the prioritized replay buffer ~\cite{prioritizedexperiencereplay} which replays transitions with a high TD training error and has been shown to speed up training.

In contrast to replay buffers, GRPO disregards diffusion trajectories after the current epoch is done. While this online sampling allows the method to train on trajectories that are generated by the most up-to-date weights, we argue that this is still highly sample inefficient and storing previous trajectories is more effective at increasing sample efficiency. 
TD error is not available in RLHF settings because rewards are provided only at the end. 
This prevents us from implementing a prioritized replay buffer. Instead, we implement a buffer that prioritizes high absolute advantage values.

\subsection{Strategy 2: Historical Hardmining}
\label{subsec:strat2}
Building on this replay buffer formulation, we propose a hard-mining strategy that prioritizes trajectories with high absolute advantage values.
Our intuition is that a trajectory with high absolute advantage, whether strongly positive or strongly negative, contains one or more timesteps that substantially changed the reward, making it highly informative training data.
We perform two different methods of trajectory replay: (1) Trajectory-level hardmining, where we find the highest $k$ trajectories based on $|A|$ in the previous few epochs, and (2) Random sampling of previous trajectories.
We find that keeping a replay buffer of only the last few epochs of trajectories was the most advantageous because sampling from an epoch too long ago resulted in out-of-distribution trajectories.

Overall, our method combines these two strategies: timestep-level reweighting and trajectory-level hard mining.
Both strategies are motivated by the idea of reweighting the training data so that more informative samples are emphasized during training. See Algorithm~\ref{Our_algorithm} for details.
\begin{figure}[htbp]
    \centering
    \includegraphics[width=\linewidth]{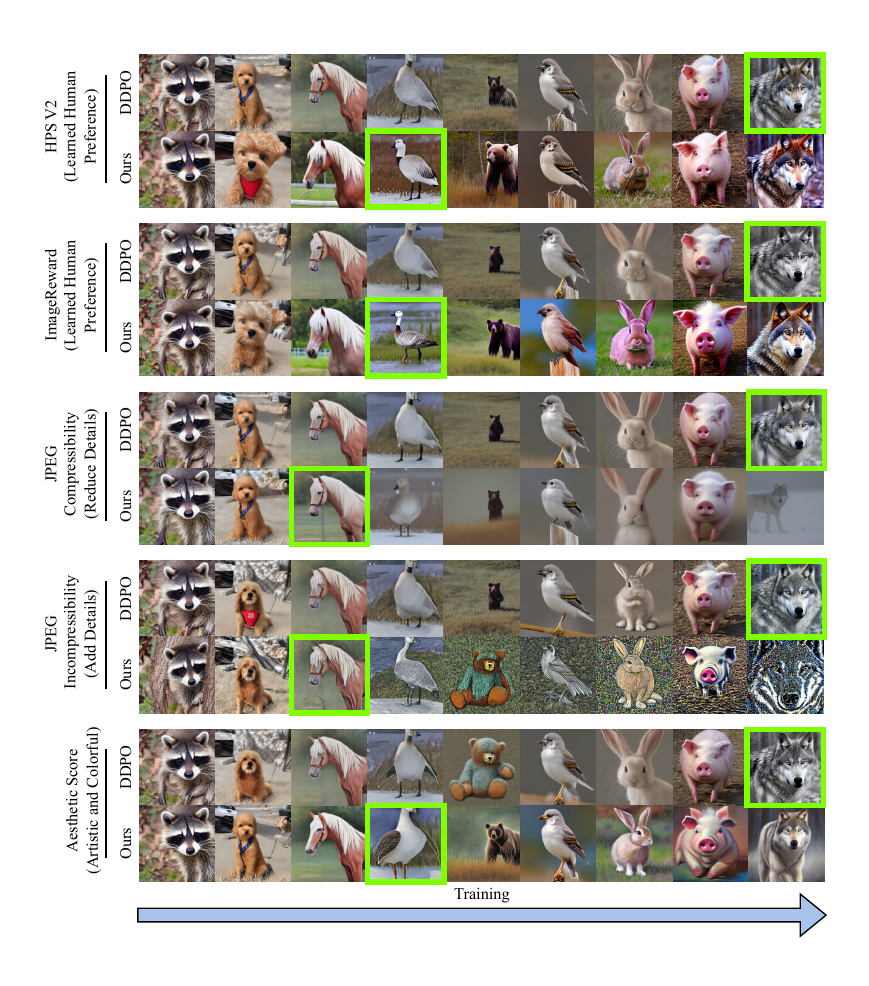}
    \vspace{-0.75cm}
    \caption{Image samples from training with the same seed. For each reward function, there are two green boxes, one for our method and one for theirs. 
     These boxes correspond to DDPO and our approach reaching the same reward level. 
     We obtain the same amount of reward in approximately half the steps in all the reward variants. 
     Note that the later images in our method's row correspond to over-optimization on the specific reward function, a phenomena common to all RLHF platforms.
        }
    \label{fig:qualitative_as_training_goes_on}
\end{figure}
\begin{center}
\resizebox{\linewidth}{!}{
\begin{minipage}{\linewidth}
\begin{algorithm}[H]
  \footnotesize
  \small
\setlength{\baselineskip}{0.9\baselineskip}
  \caption{Overall Two Strategy Algorithm}
  \begin{algorithmic}
    \STATE Initialize buffer $D=\{\}$, pretrained diffusion model $\pi$
    \FOR{$\text{epoch}=1$ \TO $\text{num\_epochs}$}
      \FOR{$i=1$ \TO $n$}
        \STATE Sample trajectory $\tau_i = (x_T, x_{T-1}, \ldots, x_0)$
      \ENDFOR
      \FOR{$i=1$ \TO $n$}
        \STATE Compute weighted loss $L_{\text{reweighted}}$ using Eq.~(2) with trajectory $\tau_i$
        \STATE $\theta = \theta - \nabla_\theta L_\text{reweighted}$
      \ENDFOR
      \IF{$D$ is not empty}
        \STATE Retrieve top $k$ samples from $D$
        \STATE Compute $L_{\text{reweighted}}$ using all top $k$ samples
        \STATE $\theta = \theta - \nabla_\theta L_\text{reweighted}$
      \ENDIF
      \STATE $D \leftarrow D \cup \{\tau_1,\ldots,\tau_n\}$
      \STATE $D \leftarrow \text{Remove\_Old\_Entries}(D)$ \# remove trajectories older than a few epochs
    \ENDFOR
  \end{algorithmic}
  \label{Our_algorithm}
\end{algorithm}
\end{minipage}
}
\end{center}
\section{Experiments}
\begin{figure*}[t!]
\centering
\begin{minipage}[t]{0.5\linewidth}
     \centering
    \captionof{table}{Generalization To Novel Prompts at 4k Reward Queries.}
    \resizebox{\linewidth}{!}{
    \begin{tabular}{@{}lcccccc@{}}
        \toprule
        & Aesthetic ($\uparrow$) & Jpeg Comp ($\uparrow$) & Jpeg Incomp ($\uparrow$) & Image Reward ($\uparrow$) & HPS v2 ($\uparrow$)\\
        \midrule
        Untrained & 5.44 & -118.75 & 118.758 & 0.48 & 0.284 \\
        DDPO  & 5.48 & -112.79 & 124.48 & 0.53 & 0.2888\\
        DDPO + Ours & \textbf{5.66} & \textbf{-44.25} & \textbf{233.28} & \textbf{0.71} & \textbf{0.3004}\\
        \bottomrule
    \end{tabular}
    }
\label{tab:Generalization_Table}
\vspace{-0.8cm}
\end{minipage}
\hfill
\begin{minipage}[t]{0.48\linewidth}
\centering
    \captionof{table}{Prompt Adherence via CLIP Score ($\uparrow$) at 4k Reward Queries }
    \resizebox{\linewidth}{!}{
    \begin{tabular}{@{}lcccccc@{}}
        \toprule
        & Aesthetic  & Jpeg Comp  & Jpeg Incomp  & Image Reward & HPS v2  \\
        \midrule
        DDPO  & 0.289 & \textbf{0.305} & 0.299 & 0.308 & \textbf{0.304}\\
        DDPO + Ours & \textbf{0.311} & 0.298 & \textbf{0.310} & \textbf{0.312} & 0.300\\
        \bottomrule
    \end{tabular}
    }
    \label{tab:CLIP_Table}
\end{minipage}
\end{figure*}

Since our method is highly applicable to existing diffusion RLHF frameworks, we choose to incorporate our method with existing baselines to show the significant increase in sample efficiency.  For all experiments, we use the same list of animal training prompts as~\cite{DDPO}. We use LoRA~\cite{LORA} weights with rank 4 matrices that were inserted into RunwayML Stable Diffusion v1.5~\cite{StableDiffusion}. For our experiments, we test our method applied to DDPO~\cite{DDPO}, DPOK~\cite{DPOK}, and B2-DiffuRL~\cite{B2-DiffuRL}. See full list of hyperparameters in the appendix.

\subsection{Reward Functions}
We test our method on 5 reward functions. We select 3 rewards from the  DDPO~\cite{DDPO} paper. We also experiment with HPS v2~\cite{wu2023human} and Image Reward~\cite{xu2023imagereward}. 
The first reward function we test on is JPEG compressibility, which prefers compressed images and favors smoother textures that are easier to compress.
Conversely, our second reward function is JPEG incompressibility, which favors highly detailed images with diverse textures.
Third, we test on the reward function given by the Aesthetic Score classifier from~\cite{schuhmann2022laion} that is trained on human aesthetic preferences. Finally, both HPS v2 and Image Reward simulate human preference by training on aesthetic and prompt-adherence human feedback on generated images.

\subsection{Sample Efficiency}
\begin{figure}[t!]
    \centering
    \includegraphics[width=\linewidth]{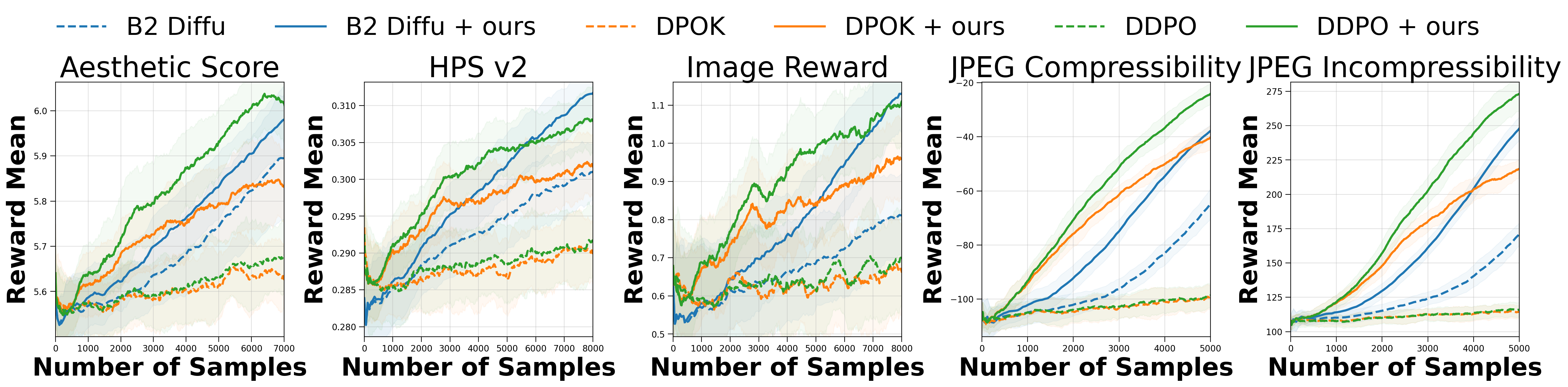}
    \vspace{-0.7cm}
    \caption{Performance of our augmentation of existing diffusion RLHF methods. As seen in the graph, our method is able to train significantly faster than simply the method alone. All results are averaged over 3 runs and are smoothed with a running average of 500 reward queries.}
    \label{fig:BaselineVsOurs}

    \vspace{1em}

    \includegraphics[width=\linewidth]{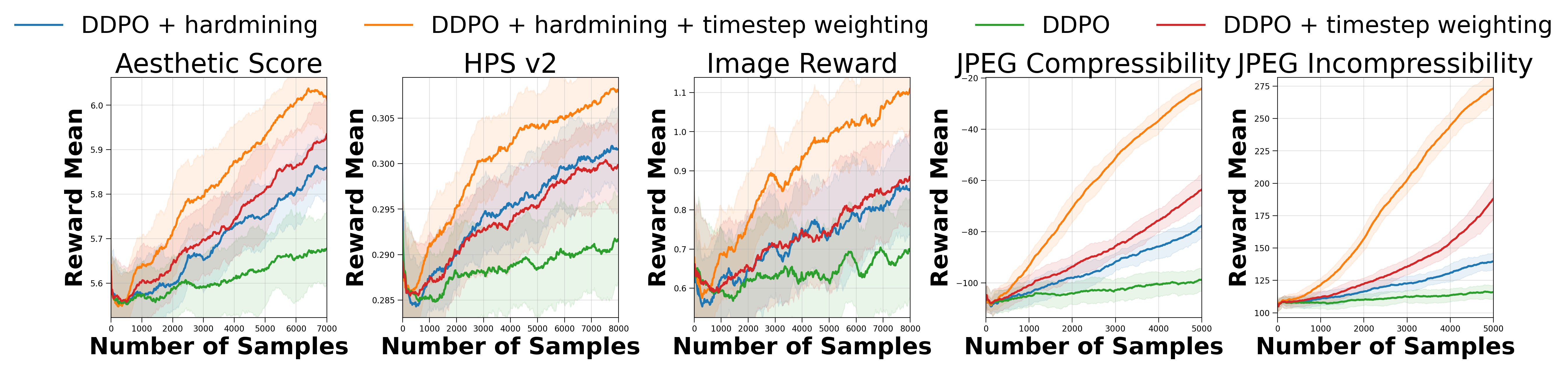}
    \vspace{-0.7cm}
    \caption{Ablation study on DDPO comparing the 4 cases: default DDPO, hardmine only DDPO, weighted timestep only DDPO, and weighted timestep and hardmined DDPO. As seen in the figure, both hardmining previous trajectories and reweighing the timesteps leads to significant increases in the performance of DDPO. Results are averaged over 3 seeds. Error margin is the std of the 3 runs. We show a running average of 500 rewards.}
    \label{fig:ablations}

    \vspace{1em}

    \includegraphics[width=\linewidth]{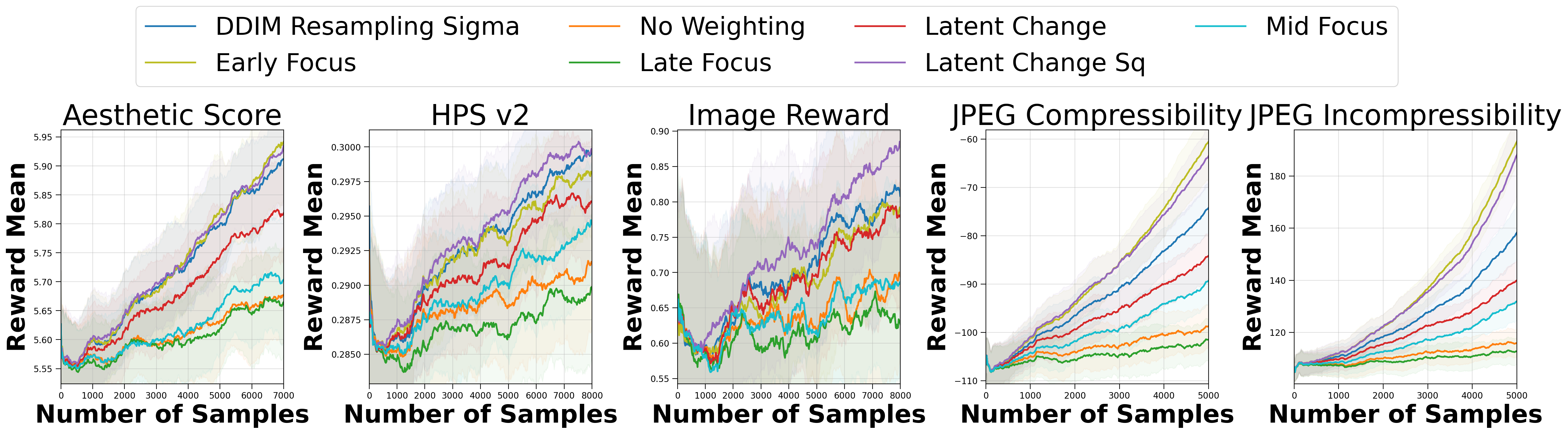}
    \vspace{-0.7cm}
    \caption{Performance of various weighting schemes. We observe that using the latent change squared gives the best results across all the rewards. In some rewards, Early focus is slightly better, but it is much inferior in HPS v2 and Image Reward. 
    }
    \label{fig:weighting_comparison}

    \vspace{1em}

    \includegraphics[width=\linewidth]{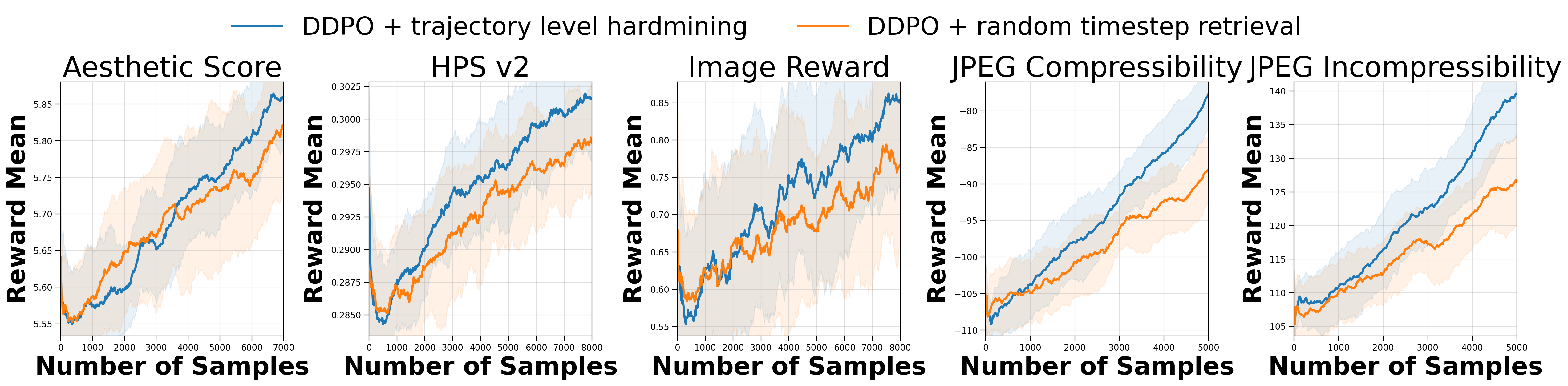}
    \vspace{-0.7cm}
    \caption{Mean reward with different retrieval methods. All results were done with timestep weighting. For all methods, we ran for 3 seeds for consistency. We show with running average of 500 rewards. All results are averaged over 3 runs.}
    \label{fig:retrieval_methods_comparison}
    \vspace{-0.50cm}
\end{figure}

In Figure \ref{fig:BaselineVsOurs}, we see that our modifications are able to achieve higher reward than all existing methods within the same budget of reward queries. For fair comparison, each baseline and its augmented counterpart share identical hyperparameters.
For qualitative examples of training images, see Figures \ref{fig:triplet_qualitative} and \ref{fig:qualitative_as_training_goes_on}.

We also show that our method is capable of generalizing to novel prompts with higher sample efficiency than existing methods. For the list of unseen prompts, we prompted ChatGPT~\cite{openai2024gpt4technicalreport} for animals not present in the prompt dataset. As seen in Table $\ref{tab:Generalization_Table}$, we find that our method still allows for more generalization than previous methods on the same budget of reward queries. 

\subsection{Ablations}
Since we do both per-timestep weighting and historical hardmining, we show that both of these methods independently make diffusion RLHF more sample efficient. We consider the 4 possibilities: (1) Neither hardmining nor timestep weighting (baseline), (2) Hardmining but not timestep weighting, (3) Timestep weighting but not hardmining, and (4) Both timestep weighting and hardmining.
Our results show that each of our techniques contributes significantly to the overall efficiency of the finetuning method. See Figure \ref{fig:ablations} where, for all the rewards, using hardmining and weighting together is more effective than using only one.

\subsection{Timestep Weighting}
Empirically, we find that timestep weighting consistently provides the most sample-efficient results, sometimes up to 6$\times$ efficiency.
Despite our reward functions focusing on different aspects of an image, our timestep weighting boosts sample efficiency for all cases, as seen in Figures \ref{fig:ablations} and \ref{fig:BaselineVsOurs}.

\noindent\textbf{Choice of Timestep Weighting.}
Using our Gaussian weighting schemes, we find that weighting earlier timesteps more heavily works better than weighting middle timesteps, which in turn works better than weighting later timesteps. This suggests that a monotonically decreasing weighting scheme is preferable. For reference, uniform (unweighted) weighting falls somewhere between the mid and early focus approaches in sample efficiency.

In our experiments, the most consistently well-performing scheme across all reward functions is the mean squared latent change, $|z_{t}-z_{t-1}|^2$ (see Fig.~\ref{fig:weighting_comparison}). 
This even outperforms the DDIM sampling standard deviation proposed in the
recent
work TempFlow~\cite{he2025tempflow}. We believe this scheme works well because it assigns higher weight to the timesteps where the model makes the largest changes to the latent representation. By prioritizing these high-change timesteps during reward-alignment training — rather than using fixed coefficients — the model is guided more effectively toward better rewards.

\subsection{Historical Hardmining}
We compare to historical sampling, where samples are randomly selected, to hardmining. As seen in Figure \ref{fig:retrieval_methods_comparison}, we see that for all reward functions, 
hardmining is more effective than simply rerunning random samples from the previous epochs.
This validates our hypothesis that large advantage trajectories are more informative than other trajectories.

\subsection{CLIP Score}
In addition to testing sample efficiency, we also test CLIP score~\cite{CLIP_Score} on the finetuned models to measure prompt-image adherence. CLIP score measures the embedding similarity between an image and its prompt, serving as a proxy for how faithfully the image reflects the text. We generated 1k images using training prompts and computed the average CLIP score, as seen in Table \ref{tab:CLIP_Table}. 
We find that there is no visible degradation of prompt adherence compared to the baseline DDPO.

\subsection{Generalization Score}
Following experiments~\cite{DDPO}, we also test generalization score. For our experiments, we first finetune a model based on a reward function for a given number of queries. Afterwards, we find a new list of prompts that were not in the list of training prompts and test how well our model keeps the reward high in those scenarios. As seen in Table \ref{tab:Generalization_Table}, we find that our method also generalizes well to other prompts. 

\section{Conclusion}
In this paper, we provide a simple but effective method to enhance the sample efficiency of diffusion RLHF. Our method is easily adaptable to existing RLHF platforms and has the potential to speed up training up to 6 times while retaining prompt-adherence and the ability to generalize to novel prompts. Our method departs from the common RLHF assumption that each timestep affects the output reward to the same extent. We ground our method on the intuition that both certain timesteps and certain trajectories are more informative than others during training. Future directions could analyze how different schedulers and $\alpha$-schedules affect the training sample efficiency or extend the method preference-based models such as D3PO~\cite{d3po}.

\bibliographystyle{splncs04}
\bibliography{main}

\cleardoublepage
\appendix
\section{Appendix}

\subsection{Hyperparameters For Baselines}
We chose to use the same hyperparameters for baselines as those chosen for all baselines in the D3PO~\cite{d3po} paper. The only difference is that we chose a smaller number samples per epoch as we found that sampling more often led to better performance. We also changed to AdamW over Adam, as consistent with most diffusion RLHF methods. Between baseline and baseline + our method, we kept all hyperparameters the same. All results were obtained with 4 RTX A5000 GPUs or 4 RTX A4000 GPUs. See a comprehensive list of hyperparameters in Tables \ref{Table:Hyparams} and \ref{Table:Hyparams_Continued}.

\subsection{Prompt List}
We used the same training prompt list as in ~\cite{DDPO}. The list of animals is listed below:
\begin{center}
    cat\;\;
    dog\;\;
    horse\;\;
    monkey\;\;
    rabbit\;\;
    zebra\;\;
    spider\;\;
    bird\;\;
    sheep\;\;
    deer\;\;
    cow\;\;
    goat\;\;
    lion\;\;
    tiger\;\;
    bear\;\;
    raccoon\;\;
    fox\;\;
    wolf\;\;
    lizard\;\;
    beetle\;\;
    ant\;\;
    butterfly\;\;
    fish\;\;
    shark\;\;
    whale\;\;
    dolphin\;\;
    squirrel\;\;
    mouse\;\;
    rat\;\;
    snake\;\;
    turtle\;\;
    frog\;\;
    chicken\;\;
    duck\;\;
    goose\;\;
    bee\;\;
    pig\;\;
    turkey\;\;
    fly\;\;
    llama\;\;
    camel\;\;
    bat\;\;
    gorilla\;\;
    hedgehog\;\;
    kangaroo\;\;
\end{center}
For generalization experiments, we used the following list of animals generated by ChatGPT:
\begin{center}
   elephant\;\;
   giraffe\;\; 
   hippopotamus\;\;  
   rhinoceros\;\;
   leopard\;\;
   cheetah\;\;
   hyena\;\;
   bison\;\;
   moose\;\;
   elk\;\;
   reindeer\;\;  
   antelope\;\;
   armadillo\;\;
   sloth\;\;
   otter\;\;
   beaver\;\;
   badger\;\;
   lynx\;\;
   bobcat\;\;
   cougar\;\;
   jaguar\;\;
   capybara\;\;
   porcupine\;\;
   platypus\;\;
   echidna\;\;
   koala\;\;
   wallaby\;\;
   wombat\;\;
   manatee\;\;
   walrus\;\;
   seal\;\;
   narwhal\;\; 
   orca\;\;
   penguin\;\; 
   albatross\;\;
   flamingo\;\;
   peacock\;\;
   owl\;\;
   eagle\;\;
   hawk\;\;
   parrot\;\;
   crocodile\;\; 
   alligator\;\;
   chameleon\;\;
   salamander\;\;
\end{center}

\subsection{Comparisons with Hardmining}
\begin{figure}
    \centering
    \includegraphics[width=0.95\linewidth]{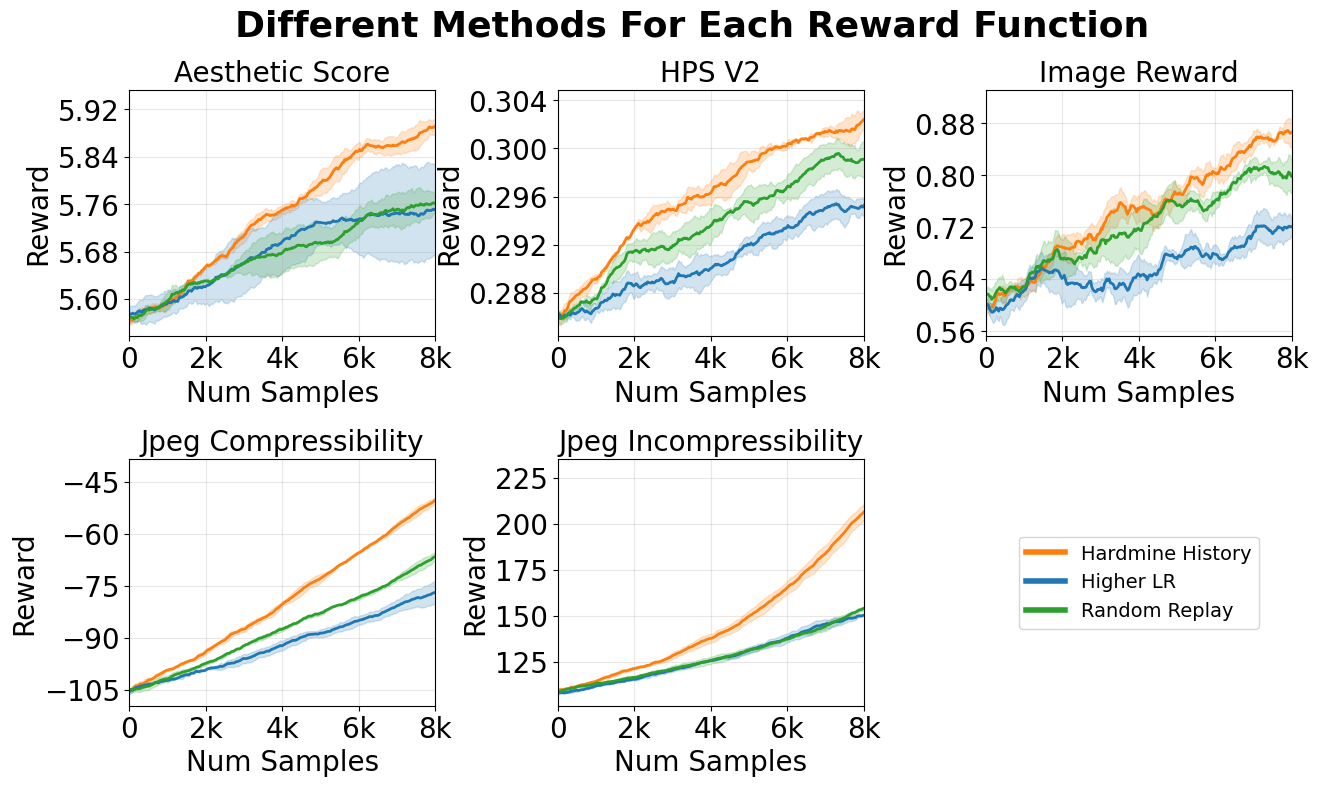}
    \caption{Comparison of Hardmining to increasing the LR and to randomly replaying samples from the current epoch. All results are averaged over 3 seeds.}
    \label{fig:appendix_hardmine_comparisons}
    \vspace{-1em}
\end{figure}

When using hardmining, even though we use lesser number of reward calls, we use more backpropagation through the network. Hence, an obvious question might be whether using hardmining helps just because there are more samples whose gradients are being backpropagrated or is hardmining helping because of the selection of more important trajectories.  
In this section, we compare hardmining with two more methods. First, we compare to simply increasing the learning rate proportional to how many more gradient steps we take. Second, we compare to randomly replaying entries from the current epoch. These two settings correspond to using faster learning rate and more gradient calls respectively, both of which are alternatives to hardmining. Note that this is different from Figure. \ref{fig:retrieval_methods_comparison} where we compared to random retrieval from previous epochs as opposed to only the current epoch. As seen in Figure \ref{fig:appendix_hardmine_comparisons}, our hardmining based on the absolute advantages does better than both increasing the learning rate and repeating samples from the current epoch, showing that hardmining helps not just because of increased samples.

\subsection{Proof That $A_k$ has mean $\displaystyle\frac{\sigma_k^2\cdot \text{std}(R)}{\sum_{i=1}^n \sigma_i^2}A_\text{final}$}
We want to find the probability distribution of $A_k$ where we assume that each $A_i$ has mean 0 and standard deviation $\sigma_i$ for $i \in \{0,1,2,...,n\}$. Furthermore, we have the condition that for a given constant $C$, $$A_0 +A_1 + A_2 +...+A_n = C$$
For simplicity of notation, assume that $k=0$. The proof can easily generalize to other values of $k$. \\\\
Let $B = A_1 + A_2+...+A_n$. The mean of $B$ is the sum of the means of $A_1+A_2,...A_n$ which is 0. The variance of $B$ is the sum of the variances of $A_1+A_2,...A_n$. This means that $\sigma_B^2 = \sigma_1^2 + \sigma_2^2 + ... + \sigma_n^2$. Finally, we have that from the summation of Gaussian distributions is still a Gaussian distribution. Thus,
$$B = A_1+A_2 + A_3 + ...+A_n \sim \mathcal{N}\left(0, \sigma_B^2\right)$$
We can rewrite our question as finding the distribution $A_0 \;|\; A_0+B=C$. We have that the for a given value $a$ such that $A_0=a$,

\begin{align}
&P(A_0=a \;|\;A_0+...+A_n=C)\nonumber\\
&= P(A_0=a \;|\;A_0+B=C) \nonumber\\
&= P(A_0=a \;\&\;A_0+B=C) / P(A_0+B=C) \nonumber\\
&= P(A_0=a \;\&\;B=C-a) / P(A_0+B=C) \nonumber\\
&\propto P(A_0=a \;\&\;B=C-a) \nonumber\\
&=  P(A_0 = a)P(B=C-a) \nonumber\\
&= \frac{1}{\sqrt{2\pi\sigma_0^2}}\exp\left({-\frac{a^2}{2\sigma_0^2}}\right)\cdot \frac{1}{\sqrt{2\pi\sigma_B^2}}\exp\left({-\frac{(C-a)^2}{2\sigma_B^2}}\right) \nonumber\\
&= \frac{1}{2\pi \sigma_0\sigma_B}\exp\left({-\frac{a^2}{2\sigma_0^2}-\frac{(C-a)^2}{2\sigma_B^2}}\right) \nonumber\\
&\propto \exp\left({-\frac{a^2}{2\sigma_0^2}-\frac{(C-a)^2}{2\sigma_B^2}}\right)\nonumber\\
&= \text{exp}\left({-\frac{a^2}{2\sigma_0^2}-\frac{(C-a)^2}{2\sigma_B^2}}\right) \nonumber\\
&= \text{exp}\left({\frac{-a^2\sigma_B^2-(C-a)^2\sigma_0^2}{2\sigma_0^2\sigma_B^2}}\right)\nonumber\\
&= \text{exp}\left({\frac{-a^2(\sigma_0^2+\sigma_B^2)+2C\sigma_0^2a - C^2\sigma_0^2}{2\sigma_0^2\sigma_B^2}}\right) \nonumber\\
&= \text{exp}\left({\frac{-a^2(\sigma_0^2+\sigma_B^2)+2C\sigma_0^2a - C^2\sigma_0^2}{2\sigma_0^2\sigma_B^2}}\right)\nonumber\\
&= \exp\Bigg(
    \frac{
        -(\sigma_0^2+\sigma_B^2)
        \left(a - C\frac{\sigma_0^2}{\sigma_0^2+\sigma_B^2}\right)^2
    }{
        2\sigma_0^2\sigma_B^2
    }
\Bigg) \nonumber\\ 
&\hspace{3.5cm}\times 
   \exp\Bigg(
    \frac{
        C^2\frac{\sigma_0^4}{\sigma_0^2+\sigma_B^2} - C^2\sigma_0^2
    }{
        2\sigma_0^2\sigma_B^2
    }
\Bigg) \nonumber\\
&= 
\underbrace{
\exp\!\left(
    \frac{
        C^2\frac{\sigma_0^4}{\sigma_0^2+\sigma_B^2}
        - C^2\sigma_0^2
    }{
        2\sigma_0^2\sigma_B^2
    }
\right)
}_{\text{constant}} \nonumber\\
&\hspace{2cm}\times 
\exp\!\left(
    \frac{
        -(\sigma_0^2+\sigma_B^2)
        \left(a - C\frac{\sigma_0^2}{\sigma_0^2+\sigma_B^2}\right)^2
    }{
        2\sigma_0^2\sigma_B^2
    }
\right)
\nonumber\\
&\propto \text{exp}\left({\displaystyle \frac{-(\sigma_0^2+\sigma_B^2)\left(a - C\frac{\sigma_0^2}{\sigma_0^2+\sigma_B^2}\right)^2}{2\sigma_0^2\sigma_B^2}}\right) \nonumber\\
&= \text{exp}\left({\displaystyle \frac{-\left(a - C\frac{\sigma_0^2}{\sigma_0^2+\sigma_B^2}\right)^2}{2\cdot \frac{\sigma_0^2\sigma_B^2}{\sigma_0^2+\sigma_B^2}}}\right)\nonumber\\
\nonumber
\end{align}

We have that the end result is a gaussian with mean 
$$\mu = C\frac{\sigma_0^2}{\sigma_0^2+\sigma_B^2} = C\frac{\sigma_0^2}{\sigma_0^2+\sigma_1^2+ ... + \sigma_n^2}$$ and variance with value  
$$\frac{\sigma_0^2\sigma_B^2}{\sigma_0^2 + \sigma_B^2} = \sigma_0^2 \left(1 - \frac{\sigma_0^2}{\sum_{i=0}^n\sigma_i^2}\right)$$
Letting $C = \text{std}(R)A_\text{final}$, we get that the distribution of $A_0$ under the assumption that $A_0 + A_1+...+A_n=\text{std}(R)A_\text{final}$ has mean $\mu = \frac{\sigma_0^2}{\sigma_0^2+\sigma_1^2 + ... + \sigma_n^2}\text{std}(R)A_\text{final}$

\begin{table*}[h!]
\centering
\caption{Hyperparameters of All Experiments}
    \begin{tabular}{|l|l|l|}
        \hline
        \textbf{Name} & \textbf{Description} & \textbf{Value} \\
        \hline
        $lr$ & learning rate & 3e-5 \\
        $optimizer$ & type of optimizer & AdamW \\
        $\xi$ & weight decay of optimizer & 1e-4 \\
        $\epsilon$ & gradient clip norm & 1.0 \\
        $\beta_1$ & $\beta_1$ of Adam & 0.9 \\
        $\beta_2$ & $\beta_2$ of Adam & 0.999 \\
        $T$ & total timesteps of inference & 20 \\
        $n$ & samples per GPU per epoch & 5 \\
        $\eta$ & eta parameter for the DDIM sampler & 1.0 \\
        $G$ & effective batch size (collectively over all gpus) & 4 \\
        $w$ & classifier-free guidance weight & 5.0 \\
        $num\_gpu$ & number of GPUs & 4 \\
        \hline
    \end{tabular}
    \label{Table:Hyparams}
\end{table*}
\begin{table*}[h!]
\centering
\caption{Additional Hyperparameters for Our Method}
    \begin{tabular}{|l|l|c|}
        \hline
        \textbf{Name} & \textbf{Description} & \textbf{Value} \\
        \hline
        $k$ & top $k$ samples to retrieve from buffer per GPU & 4 \\
        $hist\_size$ & number of previous epochs to retrieve hardmined samples & 3 \\
        $w(t)$ & per-timestep weights & Latent Change Squared\\
        \hline
    \end{tabular}
    \label{Table:Hyparams_Continued}
\end{table*}

\subsection{More Image Samples}
\begin{figure*}[h]
        \begin{center}
        \includegraphics[width=\linewidth]{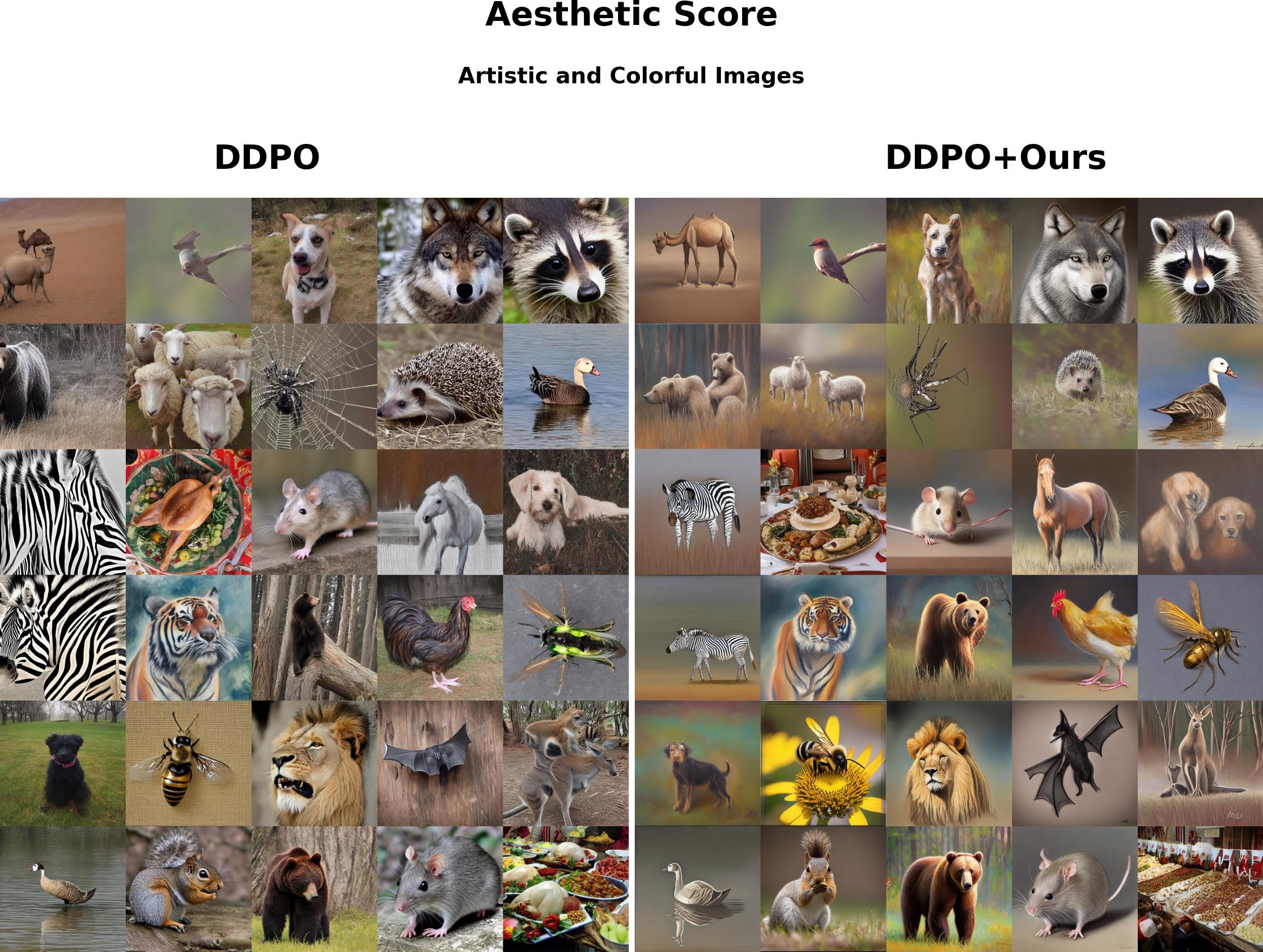}
    \end{center}
    \caption{The following is DDPO (left) vs DDPO+ours (right) after 6k reward images on Aesthetic Score. Aesthetic score mostly values bright colors, artistic value, and point of view.}
    \label{Appendix:Grid_Aesthetic_Score}
\end{figure*}

\begin{figure*}[h]
    \begin{center}
        \includegraphics[width=\linewidth]{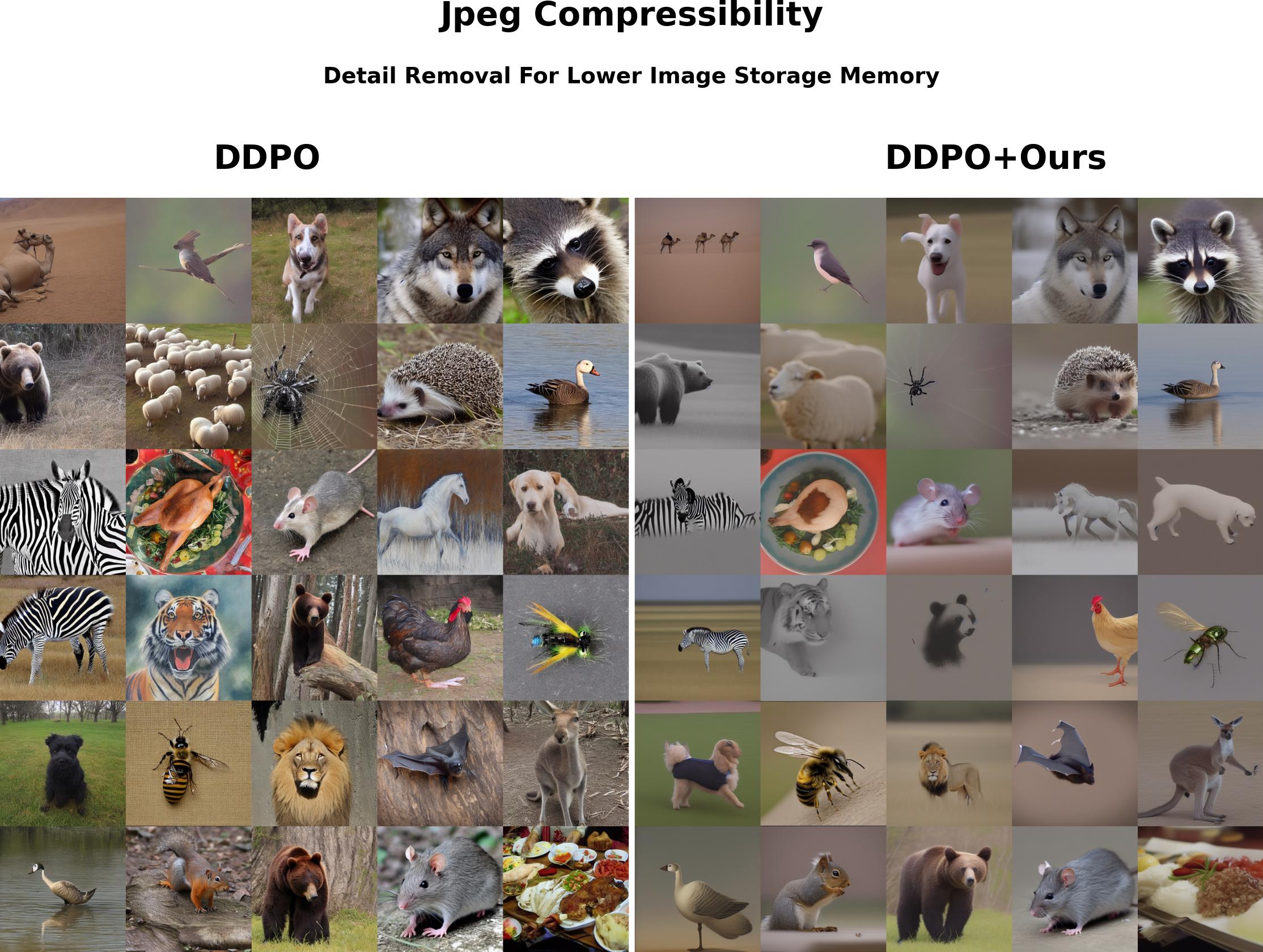}
    \end{center}
    \caption{The above is DDPO (left) vs DDPO+ours (right) after 3k reward queries on Jpeg Compression. Note that for this reward, we want the least amount of detail as possible to reduce the Jpeg memory size. Thus, the highest reward images are those with the least detail.}
    \label{Appendix:Grid_Jpeg_Comp}
\end{figure*}

\begin{figure*}[h]
    \begin{center}
        \includegraphics[width=\linewidth]{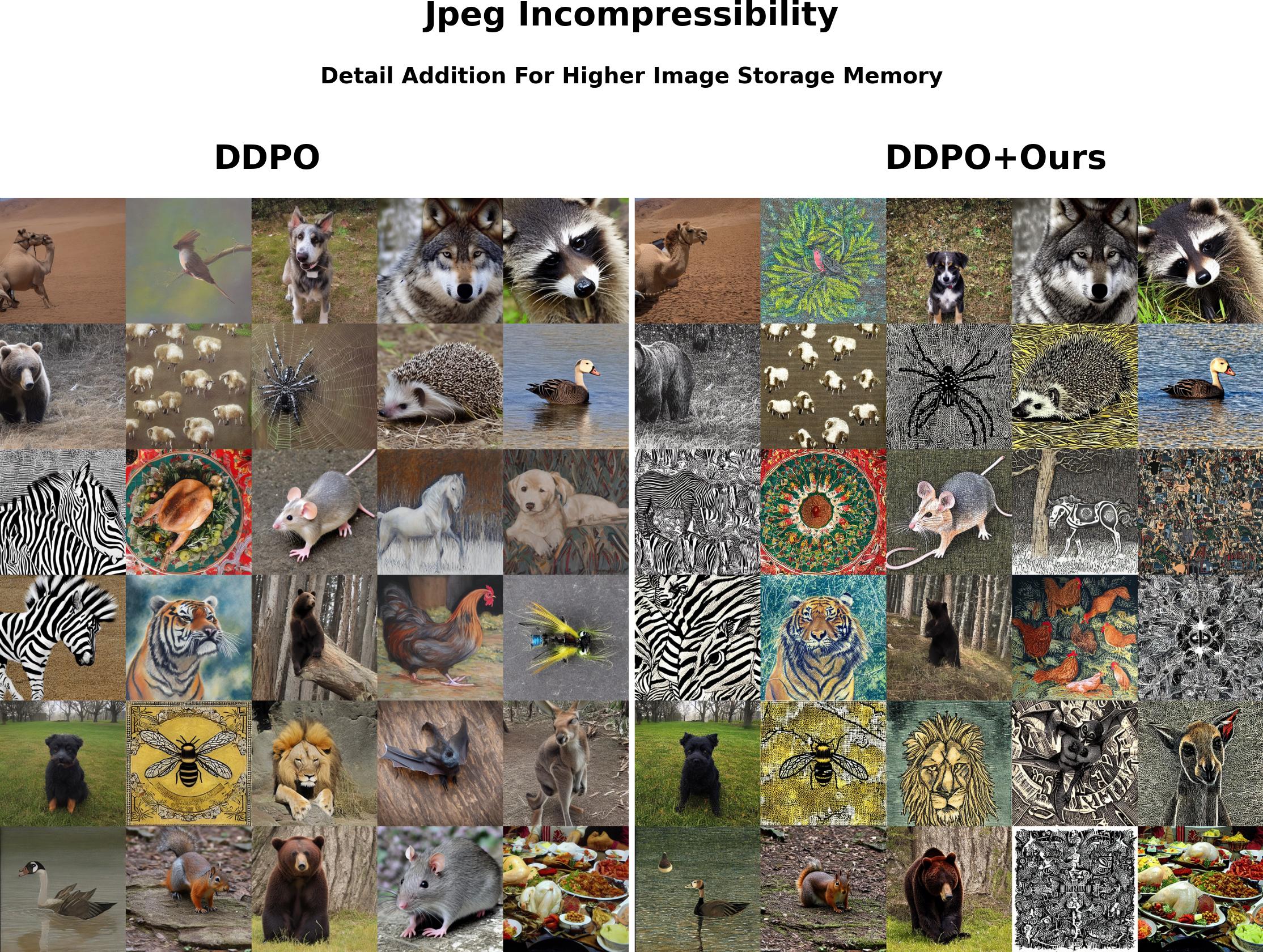}
    \end{center}
    \caption{The above is DDPO (left) vs DDPO+ours (right) after 2k reward images on Jpeg Incompression. Note that we want the most amount of detail as possible to increase the Jpeg memory size. Thus, the highest reward images are those with the most detail.}
    \label{Appendix:Grid_Jpeg_Incomp}
\end{figure*}

\begin{figure*}[h]
    \begin{center}
        \includegraphics[width=\linewidth]{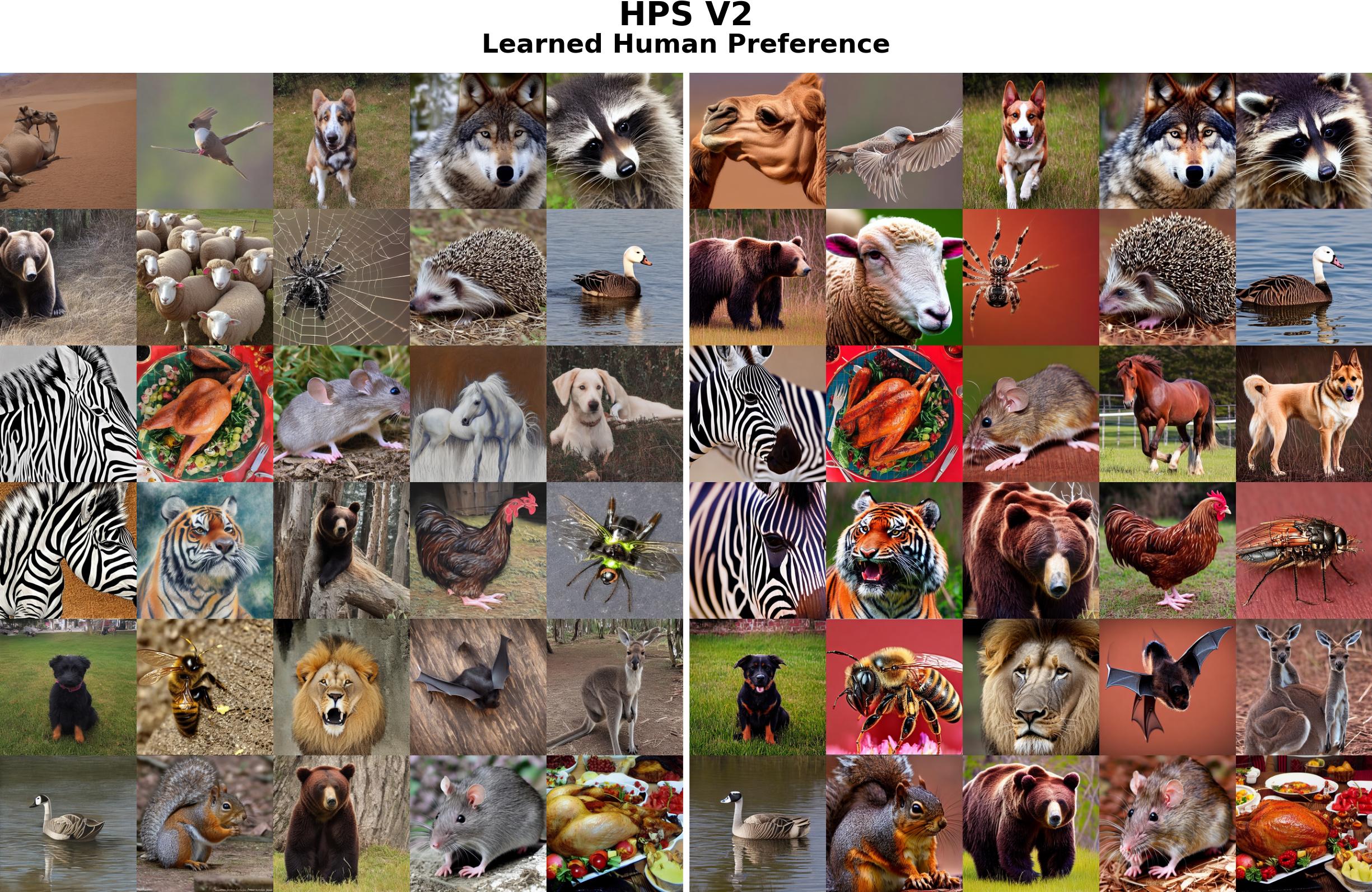}
    \end{center}
    \caption{The above is DDPO (left) vs DDPO+ours (right) after 7k reward images on HPS V2. Note that HPS V2 is trained to mimic human preference.}
    \label{Appendix:Grid_HPS_V2}
\end{figure*}

\begin{figure*}[h]
    \begin{center}
        \includegraphics[width=\linewidth]{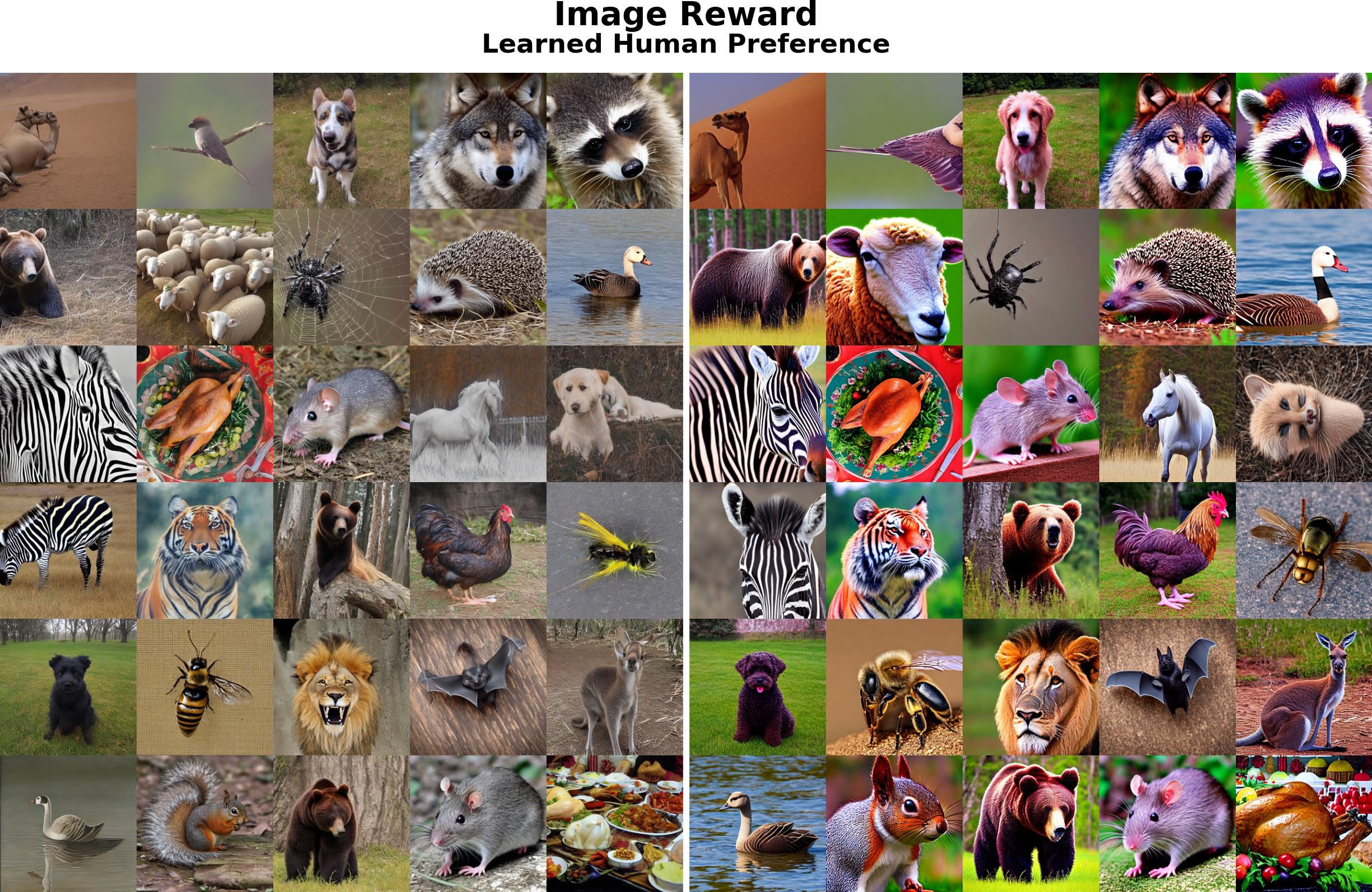}
    \end{center}
    \caption{The above is DDPO (left) vs DDPO+ours (right) after 9k reward images on Image Reward. Note that Image Reward is trained to mimic human preferences on images.}
    \label{Appendix:Grid_Image_Reward}
\end{figure*}

In figures \ref{Appendix:Grid_Aesthetic_Score}, \ref{Appendix:Grid_Jpeg_Comp}, \ref{Appendix:Grid_Jpeg_Incomp}, \ref{Appendix:Grid_HPS_V2}, and \ref{Appendix:Grid_Image_Reward}, we give a randomly selected sample of DDPO vs DDPO with our method. We use 30 random seeds in the grid for the 30 images and denoise using finetuned weights. 

\end{document}